\documentclass[sigconf,nonacm]{acmart}
\AtBeginDocument{%
  }

\setcopyright{none}
\usepackage{tabularx}
\usepackage{makecell}
\usepackage{array}
\usepackage{booktabs}
\usepackage{seqsplit}
\usepackage[utf8]{inputenc} 
\usepackage[T1]{fontenc}    
\usepackage{hyperref}       
\usepackage{url}            
\usepackage{booktabs}       
\usepackage{amsfonts}       
\usepackage{nicefrac}       
\usepackage{microtype}      
\usepackage{xcolor}         

\usepackage{amsmath}
\usepackage{graphicx}
\usepackage{multirow}
\usepackage{xcolor}
\usepackage{xcolor}

\colorlet{helpgreen}{green!50!black}
\colorlet{neutralgray}{gray}
\colorlet{hurtred}{red!70!black}
\begin{document}

\title{Knowledge or Calculator? Decomposing the Skill Premium in Verifiable Financial Agent Workflows}

\author{Jermyn Zhen Yong Bek}
\authornote{Equal contribution.}
\email{jermynbek@gmail.com}
\affiliation{%
  \institution{Independent Researcher}
  \city{Singapore}
  \country{Singapore}}

\author{Zhuang Qiang Bok}
\authornotemark[1]
\email{zq.bok@deepinsightlabs.ai}
\affiliation{%
  \institution{Deep Insight Labs}
  \city{Singapore}
  \country{Singapore}}

\author{Zhongtian Sun}
\authornotemark[1]
\authornote{Corresponding author.}
\email{Z.Sun-256@kent.ac.uk}
\affiliation{%
  \institution{University of Kent}
  \city{Canterbury}
  \country{United Kingdom}}
\affiliation{%
  \institution{University of Cambridge}
  \city{Cambridge}
  \country{United Kingdom}}

\renewcommand{\shortauthors}{Bek, Bok, and Sun}

\begin{abstract}
Financial AI agents must do more than retrieve facts: investment workflows require correct quantitative execution, reliable use of procedural resources, and auditable structured outputs. We introduce \textbf{FinSkillBench}, an evaluation suite of 2{,}603 point in time episodes across 12 subtasks in portfolio construction, risk management, and fundamental analysis, with hidden regenerable ground truth and task specific deterministic verifiers. Executing 17{,}820 episodes across 9 models and 3 resource conditions, the paired analysis across 8 models shows that curated skill packages raise mean scores by +16.2 points (0.366 to 0.528), whereas skills generated within a single episode add only +0.5 points while consuming more tokens and turns. We then decompose the curated premium by granting human authored procedural documents and executable domain tools separately: documents alone add +5.6 points, tools alone add +19.5 points, and their combination is subadditive. The premium is strongly workflow dependent: executable tools dominate numerically intensive workflows, documentation matters more when procedural or output schema guidance is the bottleneck, and interpretive tasks benefit from both. The effects are sign stable across 10 scoring variants and cluster bootstrap analyses, and an independently implemented second harness reproduces the directional pattern while showing that effect magnitudes depend on how tools and data are exposed. Overall, a measured ``skill premium'' is a property of the full model, resource, and harness system rather than of the underlying model alone. A focused composed-workflow probe reinforces this: chaining the same curated-skill subtasks into a nine-stage mandate-delivery process, an agent running on its own handoffs delivers a near-optimal but never mandate-adherent book (optimality 0.934, adherence 0/33) while every per-stage control reports success. We will release the benchmark, skill packages, verifiers, scorers, and evaluation artifacts upon publication.
\end{abstract}






\maketitle
\thispagestyle{plain}
\pagestyle{plain}

\section{Introduction}
 
Large language models are increasingly used as planning and analysis agents. In investment management, such agents are asked to solve tasks with quantitative structure: construct an optimal portfolio, monitor mandate violations, estimate stress losses, compute earnings-quality metrics, and normalize XBRL filings. These tasks differ from open-ended financial question answering: they require point-in-time discipline, structured outputs, reliable tool use, and verifiable numerical or symbolic correctness. Existing evaluations often ask whether a model knows financial concepts or can answer finance questions. That is necessary but insufficient for evaluating agentic investment workflows. A useful investment agent must know when to load domain procedures, how to pass large numeric inputs without corrupting them, which formula applies to a filing metric, and how to return a result that can be audited by downstream software. The evaluation object is therefore not only the model, but the full process by which a model uses tools, documentation, and reusable skills.

Financial AI has meanwhile moved beyond isolated prediction and question answering toward increasingly structured decision-support settings. Prior work has studied relational and geometric models for market prediction and financial news \citep{sun2023money,harit2024breaking,harit2025news}, financial recommendation \citep{sun2025ricciflowrec}, and the stability of financial NLP under semantic shift \citep{sun2025quantifying}. These developments make a broader question increasingly important: not only whether an AI model can produce a useful financial prediction, but whether a complete AI workflow remains reliable when it must retrieve information, invoke procedures and tools, and compose intermediate decisions.
 
We study this process through \textbf{FinSkillBench}, a benchmark and evaluation protocol for AI agents in investment management. The benchmark is organized around three domains: portfolio construction (mean-variance optimization, constrained optimization, tool-call parameterization, rebalancing, and Black–Litterman allocation), risk management (mandate monitoring, risk identification, stress testing, and remediation trades), and fundamental analysis (XBRL metric normalization, earnings quality, and driver decomposition).\footnote{For readers outside finance: Black–Litterman blends market-equilibrium returns with investor views; mandate monitoring checks a portfolio against client-imposed constraints; XBRL is the structured tagging format for SEC filings; the Piotroski F-Score and Beneish M-Score are published accounting screens for fundamental strength and earnings-manipulation risk.}
 
The key intervention is \emph{skill availability}. In the \textbf{no-skill} condition, the agent receives the task without curated skill documents or domain-specific tools; shared data-retrieval tools remain available. In the \textbf{curated} condition, it can discover and load human-authored skill documents and tool scripts. In the \textbf{self-generated} condition, it must first write skill documents for itself and then use them. Two further arms, \textbf{docs-only} and \textbf{tools-only}, complete a controlled decomposition that separates written procedural knowledge from executable tools (Sec.~\ref{sec:protocol}). This design evaluates not just ``Can the model solve the task?'' but ``Does a reusable skill layer improve agentic financial analysis, which component of the layer does the work, and under what assumptions?''

\paragraph{Contributions.}
Our main contribution is a controlled benchmark for evaluating how reusable procedural skills affect agentic systems in investment management:
\begin{enumerate}

\item We introduce \textbf{FinSkillBench}, a controlled benchmark for
executable financial agent workflows rather than document question
answering alone. It contains 2,603 point in time episodes across 12
subtasks in portfolio construction, risk management, and fundamental
analysis, with hidden regenerable ground truth and task specific
verifiers for numerical, constrained, ranked, and structured outputs.

\item We treat skill availability as a controlled resource intervention.
Beyond the standard comparison of no skill, curated, and self generated
conditions, docs-only and tools-only arms separate human authored
procedural documentation from executable domain tools. This reveals
that the skill premium arises from different resources across workflows,
rather than from a single knowledge effect.

\item We show that these conclusions are robust across alternative scoring
schemes and dependence aware bootstrap analyses, and examine whether
the same pattern holds in a second agent harness. The results show that
measured skill gains depend on the full model, resource, and harness
configuration, making resource and evaluation design part of the
scientific object rather than an implementation detail.

\item As a composed-process extension, we chain the same curated-skill
subtasks into a single nine-stage mandate-delivery workflow and use a
teacher-forced versus autonomous contrast to show that near-ceiling
per-subtask scores need not compose: running on its own handoffs, the agent
produces near-optimal but never mandate-adherent books while every per-stage
control reports success---a failure mode we call \emph{silent competence}.

\end{enumerate}
 
\section{Related Work}

\paragraph{Agent skills.} Anthropic introduced the Agent Skills specification as a format for equipping agents with reusable domain specific procedures~\cite{anthropic_skills2025}. SkillsBench~\cite{li2026skillsbench} evaluated this paradigm across 86 tasks and 11 general domains, finding that curated skills improve average pass rates while self generated skills provide little average benefit. We use its three condition protocol as a starting point but ask a different question: \emph{what component of a skill produces the gain in verifiable domain workflows?} FinSkillBench therefore adds a controlled decomposition of written procedural guidance and executable domain tools. Its targets are also regenerable computational objects, including convex optima, Black Litterman posteriors, stress P\&L, and structured accounting quantities, rather than human authored reference solutions alone.
 
\paragraph{Financial NLP benchmarks.} FinQA, ConvFinQA, and TAT-QA \citep{chen2021finqa,chen2022convfinqa,zhu2021tat} evaluate numerical reasoning over financial reports and tables. Recent benchmarks extend this to knowledge-intensive and document-grounded math reasoning (FinanceMath \citep{zhao2024financemath}, DocMath-Eval \citep{docmath}, FinanceReasoning \citep{financereasoning}), multimodal financial reasoning (FinMME \citep{finmme}, FinMMR \citep{tang2025mathcal}, MME-Finance \citep{mmefinance}, FinMMDocR \citep{finmmdocr}), and business-driven QA (BizFinBench \citep{bizfinbench}). Recent agentic benchmarks cover financial research \citep{bigeard2025finance}, spreadsheet-centric enterprise workflows \citep{dong2026finch}, and executable financial tool use \citep{lu2026fintoolbench}. FinSkillBench instead isolates document and tool effects through a controlled procedural-resource manipulation with point-in-time inputs, deterministic verification, and regenerable hidden ground truth.

\paragraph{Structured representations and reliability.} A complementary line of work studies how structured representations and model design affect robustness across complex AI tasks. Graph-based methods have addressed representation learning under structural heterogeneity \citep{sun2023rewiring}, uncertainty and scarce supervision \citep{sun2021generative}, and non-Euclidean or higher-order relational structure \citep{sun2025advanced,harit2025causal}. Related studies have examined behavioural change as a signal of evolving system state \citep{harit2024monitoring} and how modality-specific bias can alter model behaviour under evaluation \citep{sun2022unimodal}. These studies primarily analyse properties of individual predictive models; FinSkillBench shifts the unit of evaluation to the composed agentic process, where representations interact with procedures, tools, intermediate outputs, and the evaluation harness itself.
 
\paragraph{Tool-use and agent benchmarks.} AgentBench \citep{liu2024agentbench} and ToolBench \citep{qin2024toolllm} study multi-step tool interaction across broad domains; our benchmark narrows the domain and deepens the task contracts around investment-management methods. Classical financial methods such as mean-variance optimization \citep{markowitz1952}, Black–Litterman allocation \citep{bl1992}, factor exposure analysis \citep{ff1993}, Piotroski F-Score \citep{piotroski2000}, and Beneish M-Score \citep{beneish1999} define many of the ground-truth computations.
 
\section{Benchmark Design}
\label{sec:design}
 
\begin{figure*}[t]
\centering
\includegraphics[width=\textwidth]{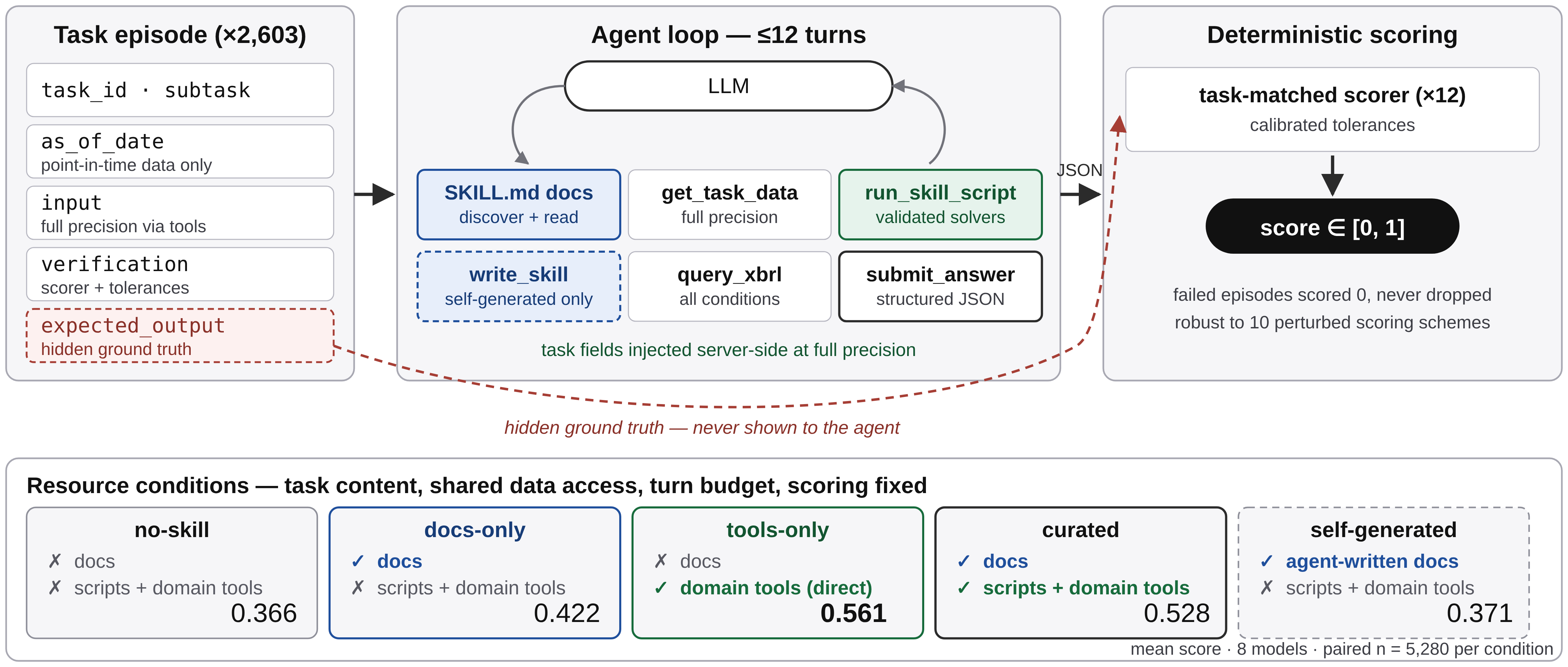}
\Description{Diagram with four panels: a task episode with point-in-time inputs and hidden expected output; the agent loop with condition-gated tools; deterministic scoring producing a score between 0 and 1; and five resource-condition cards with their mean scores.}
\caption{FinSkillBench episode anatomy. \emph{Left:} each of the 2{,}603 episodes carries point-in-time inputs, a verification spec, and hidden, deterministically derived expected output. \emph{Center:} the ReAct agent loop ($\leq$12 turns) with condition-gated resources: \texttt{SKILL.md} documents (blue) and executable scripts plus domain tools (green); shared data-retrieval tools are available in every primary-harness condition. \emph{Right:} task-matched deterministic scorers grade the submitted JSON against the hidden ground truth. \emph{Bottom:} the five resource conditions (Sec.~\ref{sec:protocol}) with overall mean scores from Sec.~\ref{sec:decomp}.}
\label{fig:anatomy}
\end{figure*}
 
\subsection{Domains and Subtasks}
 
FinSkillBench contains 12 subtasks. Figure~\ref{fig:anatomy} summarizes the episode anatomy, the agent loop, and the resource conditions. Each task episode has a unique \texttt{task\_id}, a domain skill label, a subtask, an \texttt{as\_of\_date}, an input object shown to the agent, a hidden \texttt{expected\_output}, and a \texttt{verification} object specifying the scorer. Portfolio-construction and risk-management subtasks each expose 40 tasks per subtask in the full run; fundamental-analysis subtasks have larger filtered manifests (243 / 179 / 148 for normalization / earnings quality / driver decomposition) of which the runner uses the first 100 each.
 
\subsection{Data Stack and Point-in-Time Discipline}
 
Tasks are anchored in three data substrates. Portfolio construction and risk management use a \emph{market stack} (daily prices, Fama–French factors \citep{ff1993}, macroeconomic series) from which we derive a Ledoit–Wolf shrinkage covariance \citep{lw2004}, rolling factor exposures, a composite fundamentals-plus-momentum expected-return view, and a market-cap benchmark proxy. Fundamental analysis uses a \emph{reporting stack} built from SEC EDGAR submissions and XBRL CompanyFacts. Risk management additionally relies on \emph{constructed objects}, synthetic portfolios with planted risk profiles, mandate specifications, and stress-scenario definitions, which capture institutional middle-office inputs without conflating the task with filing parsing.
 
Three design choices govern the data flow. First, all episodes carry an explicit \texttt{as\_of\_date}; system prompts forbid reliance on post-date information and tooling returns only point-in-time slices. Second, task prompts apply \emph{lossy compaction} to large numeric structures (a covariance matrix is summarized in the visible prompt, while available at full precision through tool calls). This separation measures whether agents retrieve and pipe accurate numeric input rather than reasoning from rounded text. Third, the universe is a fixed 30-name sector-stratified sample S\&P 500, sacrificing breadth for controlled and reproducible evaluation.
 
\subsection{Ground Truth and Scorers}
 
The ground truth was generated offline, stored in each episode's \texttt{expected\_output}, and never shown to the agent. Labels fall into three regimes: \emph{exact numerical} (convex-solver outputs, Black–Litterman posteriors, stress P\&L); \emph{semi-structured ranked} (risk identification, graded by ranked-list recall over a fixed label vocabulary; see Sec.~\ref{sec:evalobject}); and \emph{accounting-derived} (canonical line items, screen components, segment-level bridges with tolerances). Each subtask is paired with a verifier whose structure matches the finance-meaningful object the task produces (Table~\ref{tab:scorers}).
 
\begin{table*}[t]
\centering
\caption{Scorer-to-subtask mapping. Soft-vs-hard grading is calibrated to operational severity.}
\label{tab:scorers}
\footnotesize
\setlength{\tabcolsep}{4pt}
\renewcommand{\arraystretch}{1.15}
 
\begin{tabularx}{\textwidth}{@{}p{2.8cm}p{5.0cm}X@{}}
\toprule
\textbf{Subtask} & \textbf{Scorer} & \textbf{Object rewarded} \\
\midrule
Unconstrained optim. & \texttt{l2\_distance\_and\_objective} & Proximity to MV optimum \\
Constrained optim. &
\begin{tabular}[t]{@{}l@{}}
\texttt{constraint\_satisfaction\_and\_}\\
\texttt{objective}
\end{tabular}
& Constraint flags gate weight L2 \\
Tool-use param. & \texttt{parameter\_match} & Optimizer contract: nested constraint params \\
Rebalancing & \texttt{turnover\_compliance\_and\_objective} & Post-trade weights, turnover, trade-list consistency \\
Black--Litterman & \texttt{view\_specification\_and\_weights} & Posterior expected returns + posterior-optimal weights \\
Constraint monitoring & \texttt{exact\_match} & Mandate status + per-constraint accuracy \\
Risk identification & \texttt{ranked\_list\_recall} & Recovery + ordering of exposure categories \\
Stress testing & \texttt{absolute\_error} & Portfolio impact + sector/factor attribution \\
Risk remediation & \texttt{constraint\_satisfaction\_plus\_cost} & Feasible trades restoring compliance \\
Normalization & \texttt{metric\_absolute\_error} & Canonical line items within tolerance \\
Earnings quality & \texttt{earnings\_quality\_composite} & Piotroski components + Beneish flag + ratio terms \\
Driver decomposition & \texttt{driver\_f1\_and\_direction} & Named drivers, direction, magnitude \\
\bottomrule
\end{tabularx}
\end{table*}
 
\section{Skills as a Scientific Manipulation}
\label{sec:protocol}
 
\subsection{A Controlled Design over Procedural Resources}
 
The central manipulation is which procedural resources an agent can access.
In the primary harness, task content, shared data access, turn budget,
temperature, and scoring are held fixed; condition-specific instructions
and resource access vary only as required by the intervention.
Three main conditions define the primary comparison:
\begin{itemize}
  \item \textbf{No-skill.} The skill-access interface is present but the skill corpus is empty: discovery calls resolve to nothing. Finance domain tools (e.g., \texttt{compute\_risk\_report}, \texttt{optimize\_portfolio}) were also unavailable; the SKILL.md and scripts package is withheld. This is the reference baseline for Sec.~\ref{sec:results}. The cross-harness validation (Sec.~\ref{sec:cross-harness}) uses an analogous no-skill condition, with an additional data-access difference described in Sec.~\ref{aggregate}.
  \item \textbf{Curated.} Human-authored, domain-specific skill and tool script packages were mounted. The agent can discover and read skills, and execute tools within the same episode. 
  \item \textbf{Self-generated.} Curated skill material is withheld; the agent receives write access to a per-episode scratch space and an explicit instruction to author one to three modular skill documents \emph{before} answering, then reload them.
\end{itemize}
The design varies only the \emph{resource}: the shared interface (data retrieval and submission tools) is invariant across conditions, while access to procedural documents and executable domain tools is the manipulated factor.
 
\emph{Decomposition arms.} The three main conditions leave written skill content and executable tools confounded: no-skill withholds both, curated provides both. Two additional arms complete the decomposition on the same task IDs: \textbf{docs-only} mounts the curated \texttt{SKILL.md} documents but withholds the packages' executable scripts and the domain tools; \textbf{tools-only} exposes the domain tools with an empty skill corpus. Task content, shared data access, turn budget, temperature, and scorers are fixed across all conditions; condition specific instructions and interfaces vary only as required to expose the assigned resources. One difference in resource access is discussed in Sec.~\ref{sec:decomp}.
 
\subsection{What a Curated Skill Contains}
We define a \emph{skill} as one domain level package containing a
\texttt{SKILL.md} document and associated executable Python entry points.
FinSkillBench contains three curated skills, one each for portfolio
construction, risk management, and fundamental analysis, covering five,
four, and three subtasks, respectively. The packages were authored
specifically for FinSkillBench by the benchmark developers. 
Each package combines \textbf{declarative} content (Markdown procedures, formulas, common failure modes, API usage patterns) and \textbf{imperative} content (small validated Python entry points invoked through a constrained script-execution tool with JSON I/O). The split addresses a failure mode observed in piloting: language models reliably invent the shape of a computation but fail to pass large numeric structures through tool calls at full precision. For example, a 900-float covariance matrix emitted by a model and re-parsed into an optimizer differs from the source at the last few digits, and those differences change the optimum. The curated design therefore keeps numeric-heavy routines inside validated scripts, with server-side injection of full-precision task fields (no round-trip through model-generated JSON).

\subsection{Agent Loop}
 
The agent follows a function-calling, ReAct-style loop with a common task
interface and shared retrieval and submission tools. Condition-specific
interfaces expose skill discovery and loading, skill authoring
(self-generated only), and executable domain tools according to the assigned
resource condition. Shared tools retrieve full task fields and domain data,
while curated script execution is scoped to the active skill directory.
Turn budget is capped at 12 with up to 3 retries for transient provider
failures. Episodes that fail to emit valid JSON through the submission tool
are scored \texttt{invalid\_submission} or
\texttt{incomplete\_submission} rather than silently dropped.

\section{Experimental Setup}
\label{sec:setup}

\textbf{Models.} We evaluated 9 models spanning frontier and open-weight families: \texttt{gpt-5.4}, \texttt{claude-sonnet-4.6}, \texttt{gpt-4.1}, \texttt{gemini-2.5-pro}, \texttt{grok-4}, \texttt{DeepSeek-V3.2}, \texttt{Phi-4}, \texttt{glm-5.1}, \texttt{gemini-3.1-flash-lite}.
 
\textbf{Run.} Full coverage: 9 models $\times$ 3 conditions $\times$ 12 subtasks. Tasks per cell: 40 for PC and RM subtasks, 100 for FA subtasks. One sample per cell. Total: \textbf{17{,}820 episodes}. Execution: 12 concurrent OS processes (one per subtask runner), 16 \texttt{ThreadPoolExecutor} workers each. Temperature 0.7; max turns 12; eval retries 3.
 
\textbf{Validity.} 17{,}176 / 17{,}820 evaluations (96.4\%) returned valid JSON. 644 (3.61\%) were \texttt{invalid\_submission}; 3{,}736 (21.0\%) hit the turn budget (\texttt{max\_turns\_exhausted}); 75 (0.4\%) were incomplete\_submission. 
Phi-4 exhausted the turn budget on all 1{,}980 episodes and scored 0.000.
Unless stated otherwise, aggregate, domain, subtask, cost, and decomposition
analyses therefore exclude Phi-4; Table~\ref{tab:models} retains it for
completeness.

\section{Results}
\label{sec:results}
 
\subsection{Aggregate Skill Effect}
We report condition-level performance under the
analysis set defined in Sec.~\ref{sec:setup} in Table~\ref{tab:aggregate}. Curated skills give a large,
stable improvement over no-skill, while self-generated skills barely move
the needle.
 
\begin{table}[ht]
\centering
\small
\caption{Aggregate condition means (5{,}280 evaluations per condition). Confidence intervals are bootstrap 95\% intervals (10{,}000 resamples) on the paired condition-vs-no-skill deltas.}
\label{tab:aggregate}
\begin{tabular}{@{}lccc@{}}
\toprule
\textbf{Condition} & \textbf{Mean Score} & \textbf{$\Delta$ vs.\ No-Skill} & \textbf{95\% CI} \\
\midrule
No-Skill & 0.366 & --- & --- \\
\textbf{Curated} & \textbf{0.528} & \textcolor{helpgreen}{\textbf{+0.162}} & [+0.152, +0.171] \\
Self-Generated & 0.371 & \textcolor{neutralgray}{+0.005} & [$-$0.002, +0.011] \\
\bottomrule
\end{tabular}
\end{table}
 
\subsection{Domain-Level Effects}
 
Curated-skill gains are largest where the skill provides executable computational primitives.
 
\begin{table}[ht]
\centering
\small
\caption{Domain-level results. PC and RM benefit most because curated skills contain validated solvers and precise input-passing instructions. CI shown for the curated $-$ no-skill paired delta.}
\label{tab:domain}
\resizebox{\columnwidth}{!}{%
\begin{tabular}{@{}lccccc@{}}
\toprule
\textbf{Domain} & \textbf{No-Skill} & \textbf{Curated} & \textbf{Self-Gen} & \textbf{Curated $\Delta$} & \textbf{Self-Gen $\Delta$} \\
\midrule
Portfolio construction & 0.307 & \textbf{0.585} & 0.301 & \textcolor{helpgreen}{\textbf{+0.278}} [+0.258, +0.298] & $-$0.006 [$-$0.015, +0.004] \\
Risk management & 0.269 & \textbf{0.486} & 0.303 & \textcolor{helpgreen}{\textbf{+0.218}} [+0.195, +0.239] & +0.034 [+0.018, +0.049] \\
Fundamental analysis & 0.458 & \textbf{0.512} & 0.454 & \textcolor{helpgreen}{\textbf{+0.054}} [+0.045, +0.064] & $-$0.004 [$-$0.014, +0.006] \\
\bottomrule
\end{tabular}}
\end{table}
 
\subsection{Subtask-Level Effects}
 
The picture sharpens at subtask resolution (Table~\ref{tab:subtask}). The largest improvements occur on subtasks where the no-skill baseline is near zero or modest; subtasks with already-high baselines show small or negligible gains.
 
\begin{table}[ht]
\centering
\small
\caption{Subtask-level results. Sorted by Curated $-$ No-Skill $\Delta$.}
\label{tab:subtask}
\resizebox{\columnwidth}{!}{%
\begin{tabular}{@{}llcccc@{}}
\toprule
\textbf{Subtask} & \textbf{Domain} & \textbf{No-Skill} & \textbf{Curated} & \textbf{Self-Gen} & \textbf{$\Delta$} \\
\midrule
Unconstrained optimization & PC & 0.247 & 0.719 & 0.236 & \textcolor{helpgreen}{\textbf{+0.472}} \\
Constrained optimization & PC & 0.148 & 0.505 & 0.127 & \textcolor{helpgreen}{\textbf{+0.358}} \\
Risk identification & RM & 0.023 & 0.332 & 0.055 & \textcolor{helpgreen}{\textbf{+0.309}} \\
Rebalancing & PC & 0.183 & 0.451 & 0.164 & \textcolor{helpgreen}{+0.268} \\
Stress testing & RM & 0.017 & 0.281 & 0.027 & \textcolor{helpgreen}{+0.265} \\
Black–Litterman & PC & 0.153 & 0.375 & 0.147 & \textcolor{helpgreen}{+0.222} \\
Constraint monitoring & RM & 0.597 & 0.775 & 0.651 & \textcolor{helpgreen}{+0.178} \\
Risk remediation & RM & 0.438 & 0.558 & 0.478 & \textcolor{helpgreen}{+0.120} \\
Earnings quality & FA & 0.436 & 0.535 & 0.474 & \textcolor{helpgreen}{+0.099} \\
Tool-use parameterization & PC & 0.803 & 0.872 & 0.831 & \textcolor{helpgreen}{+0.069} \\
Driver decomposition & FA & 0.256 & 0.322 & 0.249 & \textcolor{helpgreen}{+0.066} \\
Normalization & FA & 0.681 & 0.678 & 0.638 & \textcolor{neutralgray}{$-$0.003} \\
\bottomrule
\end{tabular}}
\end{table}
 
Tool-use parameterization has high no-skill performance because the expected output is a structured parameter translation rather than a full computation. Normalization shows no curated benefit because the base task is largely direct XBRL extraction plus simple arithmetic, already accessible via the global XBRL query tool.
 
\subsection{Model-Level Effects}
\label{sec:modeleffects}
Skill effects are heterogeneous across base models.
\texttt{claude-sonnet-4.6} starts from the strongest no skill baseline
(0.534), while \texttt{gpt-4.1} shows the largest raw curated difference
(+0.405) from the weakest no skill baseline among the six models without
severe execution failures (0.330). A trace-level audit of this run cautions against reading the per-model deltas as a capability ranking. A start-of-run scheduling artifact caused 61\% of \texttt{gpt-5.4}'s curated skill-script executions to record spurious 30-second timeouts (under 1\% for every other model; zero of 66 replayed timeouts reproduce), depressing its curated cells on 8 of 12 subtasks, and three scorer definitions penalize verbose responders most (Sec.~\ref{sec:evalobject}).
 
\begin{table}[ht]
\centering
\small
\caption{Per-model overall and condition means. ``Overall'' is the mean over all 1{,}980 episodes per model across the three conditions (FA subtasks carry 100 tasks each vs.\ 40 for PC/RM). Read as bands rather than a strict ordering (Sec.~\ref{sec:modeleffects}).}
\label{tab:models}
\resizebox{\columnwidth}{!}{%
\begin{tabular}{@{}lccccc@{}}
\toprule
\textbf{Model} & \textbf{Overall} & \textbf{No-Skill} & \textbf{Curated} & \textbf{Self-Gen} & \textbf{Curated $\Delta$} \\
\midrule
gpt-4.1 & 0.488 & 0.330 & \textbf{0.735} & 0.398 & \textcolor{helpgreen}{\textbf{+0.405}} \\
gemini-2.5-pro & 0.490 & 0.380 & \textbf{0.678} & 0.412 & \textcolor{helpgreen}{\textbf{+0.298}} \\
DeepSeek-V3.2 & 0.521 & 0.458 & \textbf{0.658} & 0.447 & \textcolor{helpgreen}{+0.200} \\
gemini-3.1-flash-lite & 0.487 & 0.419 & \textbf{0.609} & 0.434 & \textcolor{helpgreen}{+0.191} \\
claude-sonnet-4.6 & 0.560 & 0.534 & \textbf{0.659} & 0.488 & \textcolor{helpgreen}{+0.125} \\
gpt-5.4 & 0.452 & 0.430 & \textbf{0.520} & 0.405 & \textcolor{helpgreen}{+0.091} \\
glm-5.1 & 0.268 & 0.274 & 0.285 & 0.245 & \textcolor{neutralgray}{+0.011} \\
grok-4 & 0.107 & 0.104 & 0.077 & \textbf{0.138} & \textcolor{hurtred}{$-$0.027} \\
Phi-4 & 0.000 & 0.000 & 0.000 & 0.000 & \textcolor{neutralgray}{0.000} \\
\bottomrule
\end{tabular}}
\end{table}
 
In a sensitivity re-analysis that repairs all four issues, \texttt{gpt-5.4}'s curated mean rises from 0.520 to 0.680, and the top six curated scores fall within a band of 0.10 spanning frontier, mid, and economical model tiers. We therefore read Table~\ref{tab:models} as broad bands rather than a
capability ranking; the corrected curated scores do not follow a
monotonic ordering by model tier, and we draw no conclusion about which
model benefits most. Two execution failure patterns limit effective skill use. \texttt{Phi-4} and \texttt{grok-4} exhaust the turn budget on most episodes (100\% and 64 to 88\% by condition), while \texttt{glm-5.1} emits malformed tool call arguments on 37.9\% of its curated condition calls.
 
\subsection{Cost and Interaction Overhead}
 
Self-generated skills are expensive without delivering accuracy gains: median tokens per episode rise from 19{,}952 (no-skill) and 22{,}546 (curated) to 40{,}034 (self-generated), median latency from 59.8\,s and 49.8\,s to 76.8\,s, and mean turns from 7.5 and 7.3 to 9.3. Self-generated skills load in 97.7\% of non-Phi-4 episodes, but the generated skills rarely improve accuracy. ``Writing down a procedure'' inside the episode is not equivalent to having a validated domain skill with tested scripts and precise data-access guidance.
 
\subsection{Knowledge or Calculator? Decomposing the Premium}
\label{sec:decomp}
 
Is the skill premium the \emph{knowledge} (which procedure applies) or the \emph{calculator} (validated executable numerics)? We evaluate the two decomposition arms of Sec.~\ref{sec:protocol} on the same 8 models and identical task IDs as the main conditions (paired, $n{=}5{,}280$ episodes per condition). Table~\ref{tab:decomp} reports episode-pooled means.
 
\begin{table}[ht]
\centering
\small
\caption{Five-condition decomposition on identical task IDs (8 models, paired $n{=}5{,}280$ per condition). Self-generated overall mean: 0.371. Bold marks the best condition per row.}
\label{tab:decomp}
\setlength{\tabcolsep}{4.5pt}
\begin{tabular}{@{}lcccc@{}}
\toprule
 & \textbf{No-skill} & \textbf{Docs-only} & \textbf{Tools-only} & \textbf{Curated} \\
\midrule
Overall & 0.366 & 0.422 & \textbf{0.561} & 0.528 \\
\midrule
Risk identification & 0.023 & \textbf{0.382} & 0.029 & 0.332 \\
Stress testing & 0.017 & 0.086 & \textbf{0.469} & 0.281 \\
Constrained optim. & 0.148 & 0.176 & \textbf{0.629} & 0.505 \\
\bottomrule
\end{tabular}
\end{table}
 
Overall, the premium splits into a documents effect of $+0.056$, a tools effect of $+0.195$, and an interaction of $-0.089$. Three findings structure it by investment-management workflow. \emph{First}, knowledge-bound middle-office tasks are carried by documents: on risk identification, docs-only scores 0.382, the highest of all conditions, while tools-only reaches only 0.029. \emph{Second}, computational tasks are carried by tools: on stress testing, tools-only scores 0.469 against 0.086 for docs-only, with the same pattern on constrained optimization (0.629 vs.\ 0.176), rebalancing, and Black-Litterman. \emph{Third}, interpretive fundamental-analysis tasks need both: the combined package is best on earnings quality (0.535) and driver decomposition (0.322), while normalization stays flat across the no-skill, docs, tools, and curated conditions (0.676-0.681) because the global XBRL tool already covers it. A trace audit qualifies the risk-identification result: its verifier matches a four-label vocabulary that the RM \texttt{SKILL.md} states verbatim, and in the cross-harness arms (Sec.~\ref{sec:cross-harness}) the share of predicted labels outside that vocabulary falls from 85\% to 34\% when the document is present. Thus, part of the document effect there is supplying the output ontology the verifier scores against, not only analytical procedure (cf.\ Sec.~\ref{sec:evalobject}).

The negative interaction reflects a real cost. Under the fixed 12 turn budget, adding documents to tools reduces scores on several computation subtasks, including stress testing from 0.469 to 0.281, because discovery and reading compete with execution turns. Stacking procedural resources is therefore not free. The decomposition also sharpens the self generation result. The docs only and self generated conditions both lack executable scripts and share the global XBRL tool, yet score 0.422 versus 0.371 overall and 0.382 versus 0.055 on risk identification. This difference is consistent with document quality and in episode authoring cost rather than executable tool access. The tools only condition exposes domain tools directly, whereas the curated condition routes them through \texttt{run\_skill\_script}. The comparison between these conditions therefore also includes a difference in how tools are accessed, consistent with the cross harness results in Sec.~\ref{sec:cross-harness}.
 
\subsection{Scorer Robustness}
\label{sec:robust}
 
To test whether these conclusions depend on the hand-set tolerances and weights behind Table~\ref{tab:scorers}, we re-scored all 17{,}820 episodes offline under ten schemes varying three factors independently: every tolerance halved and doubled; the error-to-score map replaced by an unclipped smooth map and by a hard step map with no partial credit; and all composite weights set equal, including removal of the constrained-optimization gate. The curated effect spans $+0.138$ to $+0.161$ across all ten schemes and the self-generated effect $+0.004$ to $+0.008$; all twelve per-subtask curated deltas keep their sign in every scheme (eleven positive, normalization negative). Per-episode Spearman $\rho \geq 0.921$ throughout, and Kendall $\tau$ on model ordering is 0.778 to 1.0, with a maximum displacement of two positions confined to three models separated by 0.002. Exactly one episode of 17{,}820 exceeded a score of 1.0 (at 1.0461, the ranked-list clamping case), fixed in the same re-scoring pass.
 
\section{Cross-Harness Validation}
\label{sec:cross-harness}
 
To test whether the no-skill vs.\ curated effect is robust to agent-implementation choices, we replicate the comparison on an independent harness: the Nous Research \textbf{Hermes Agent} \citep{nousresearch2026hermesagent} with the same domain skills, tools, prompt template, submission protocol, and turn budget (max\_turns~$=$~12); only the Hermes \texttt{skills\_list} and \texttt{skill\_view} tools were activated, with all other built-in skills, toolsets, memory and personality off. Eight models (gpt-5.4, gpt-4.1, claude-sonnet-4.6, gemini-2.5-pro, glm-5.1, grok-4.20, DeepSeek-V3.2, gemma-4-31b-it) were evaluated under both conditions across all 12 subtasks: the no-skill run sampled the first 20 episodes per subtask (240 per model, 60 FA / 100 PC / 80 RM; \textbf{1{,}920} total), while the curated run used the full 660 episodes per model (\textbf{5{,}280} total). Both runs completed with zero runner errors and used the same task data and scorers as the main study. Because episode coverage differs between conditions, the cross-condition
mean differences below are descriptive rather than matched
treatment-effect estimates.
 
\subsection{Aggregate and Domain Effects} 
\label{aggregate}
 
\begin{table}[ht]
\centering
\small
\caption{Hermes harness: descriptive aggregate, domain, and model means.
The no skill run contains 1{,}920 episodes and the curated run 5{,}280,
so cross condition differences are not matched treatment effect estimates.
Separate 95\% CIs are shown for each condition. Model identifiers follow
provider API naming; exact route strings will be released with the artifact.}
\label{tab:hermes}

\resizebox{\columnwidth}{!}{%
\begin{tabular}{@{}lccccc@{}}
\toprule
\textbf{Group} &
\textbf{$N$} &
\textbf{No Skill (95\% CI)} &
\textbf{$N$} &
\textbf{Curated (95\% CI)} &
\textbf{Raw $\Delta$} \\
\midrule

\multicolumn{6}{l}{\emph{Overall}} \\
All &
1920 &
0.354 [0.338, 0.370] &
5280 &
0.679 [0.669, 0.688] &
+0.325 \\

\midrule
\multicolumn{6}{l}{\emph{By domain}} \\
Fundamental analysis &
480 &
0.439 [0.408, 0.470] &
2400 &
0.567 [0.553, 0.582] &
+0.129 \\

Portfolio construction &
800 &
0.325 [0.301, 0.350] &
1600 &
0.760 [0.744, 0.776] &
+0.435 \\

Risk management &
640 &
0.325 [0.297, 0.354] &
1280 &
0.785 [0.767, 0.802] &
+0.459 \\

\midrule
\multicolumn{6}{l}{\emph{By model}} \\
claude-sonnet-4.6 &
240 &
0.304 [0.257, 0.353] &
660 &
0.795 [0.775, 0.815] &
+0.490 \\

DeepSeek-V3.2 &
240 &
0.371 [0.325, 0.419] &
660 &
0.686 [0.659, 0.712] &
+0.315 \\

gemini-2.5-pro &
240 &
0.333 [0.291, 0.375] &
660 &
0.477 [0.441, 0.512] &
+0.144 \\

gemma-4-31b-it &
240 &
0.375 [0.332, 0.418] &
660 &
0.642 [0.614, 0.670] &
+0.267 \\

glm-5.1 &
240 &
0.397 [0.350, 0.444] &
660 &
0.779 [0.758, 0.801] &
+0.383 \\

gpt-4.1 &
240 &
0.285 [0.244, 0.327] &
660 &
0.699 [0.672, 0.726] &
+0.414 \\

gpt-5.4 &
240 &
0.408 [0.365, 0.451] &
660 &
0.660 [0.631, 0.687] &
+0.252 \\

grok-4.20 &
240 &
0.357 [0.314, 0.401] &
660 &
0.691 [0.664, 0.719] &
+0.334 \\

\bottomrule
\end{tabular}}
\end{table}
 
The Hermes harness shows the same directional pattern across all three
domains: curated-condition means exceed no-skill means. The raw overall
difference is +0.325 (0.679 vs.\ 0.354), but because the two runs use
different episode coverage, we treat this and the domain- and model-level
differences as descriptive. One factor that may contribute to the larger
raw difference is the Hermes access design: tools are loaded as Python
functions, and withholding them in the no-skill condition also removes
the point-in-time reference data they carry.
 
\paragraph{Note on no-skill design across runs.}
Unlike the primary harness, where shared data retrieval remains available
in every condition, withholding domain tools in Hermes also removes the
reference data carried by those tools. The raw Hermes condition difference
is $+32.5$ pp, roughly twice the paired $+16.2$ pp effect in the primary
harness. However, the Hermes difference combines changes in skill documents,
executable tool access, and data access, and the two conditions use different
episode coverage. The comparison is therefore descriptive and does not isolate
a harness effect. The decomposition of documents, tools, and their interaction
in the primary harness is reported in Sec.~\ref{sec:decomp}. 
 
At the model level, claude-sonnet-4.6 shows the largest raw mean
difference (+0.490). Open-weight models also show large descriptive
differences (glm-5.1 +38.3\,pp, DeepSeek-V3.2 +31.5\,pp, gemma-4-31b-it
+26.7\,pp). Under the curated condition, glm-5.1 reaches the
second-highest mean score (0.779), showing strong performance from a
locally deployable model in this harness.
 
\subsection{Cross-Harness Comparison: Same Pattern, Different Magnitude}
 Raw subtask-level condition differences also vary across harnesses. Risk
identification changes from 0.023 to 0.332 in FinSkillBench and from
0.043 to 0.457 in Hermes, while constrained optimization shows different
raw differences despite similar no-skill means (0.148 vs.\ 0.177).
Because the Hermes conditions use different episode coverage, these
comparisons are descriptive and do not isolate a harness effect. They are
nevertheless consistent with the broader conclusion that measured skill
differences depend on how a harness exposes tools and data.
 
\subsection{Implications}
Two practical implications follow. Claims of ``$X$\,pp skill uplift''
from a single benchmark are conditional on its tool and data access
design, condition instructions, turn budget, no-skill baseline, and
episode coverage. We therefore recommend reporting these quantities
alongside the harness configuration rather than as a single transferable
number.
 
\section{Discussion}
\subsection{What Curated Skills Measure}
Curated skill gains are not pure knowledge gains: skills combine
procedural guidance, tool-use instructions, executable scripts, and
failure-mode warnings, and Sec.~\ref{sec:decomp} quantifies how the
credit splits. The pattern is specific to investment workflows. Portfolio
construction and stress testing depend on precise point-in-time numerics
and therefore favor executable tools, while risk identification depends
more on risk taxonomy and output ontology and therefore favors documents.
Interpretive fundamental-analysis tasks benefit from both, while
normalization gains little because shared XBRL retrieval already covers
much of the workflow. Thus, the value of a financial skill depends on
whether the bottleneck is procedural categorization, numerical execution,
or interpretation.
 
\subsection{Why Self-Generated Skills Underperform}
 
The self-generated condition supports a specific claim: within a 12-turn episode, unvalidated agent-written documents -- authored, reloaded, and applied under the same budget -- do not match pre-tested documents plus executable solvers. The same-access comparison in Sec.~\ref{sec:decomp} (docs-only 0.422 vs.\ self-generated 0.371, with equal script access) locates the gap in document quality and authoring cost rather than tool access. A smaller follow-up experiment (4 models, 384 paired cells, document truncation removed) refines the null: self-generation there shows a small positive effect over no skill ($+0.039$ [$+0.015$, $+0.064$]), and capability-matched human-authored documents beat agent-authored ones by $+0.066$ [$+0.043$, $+0.090$]. Self-generated skills may also still help in a multi-episode setting where skills persist, are tested, and are corrected over time; this benchmark does not evaluate that scenario.
 
\subsection{Evaluation as a Scientific Object}
\label{sec:evalobject}
 
The benchmark demonstrates that conclusions about ``financial agent capability'' depend strongly on evaluation design. For example: if invalid and max-turn submissions are dropped, score estimates are inflated relative to deployment-like reliability, and if no-skill agents can call the same deterministic solvers as curated agents, the measured skill effect is confounded. A post-hoc trace audit of the full run makes this concrete. The scoring pipeline reproduces stored scores exactly (135 sampled episodes, zero mismatches), but three metric definitions partly scored presentation rather than analysis: stress testing compared answers against fractional ground truth while the prompt's input data used whole-number percent with no declared convention, risk identification matched labels by exact string equality, and \texttt{parameter\_match} counted unrequested extra keys against the score. Correcting all three identically for every model moves the pooled curated effect over the six models without severe
execution failures from $+0.218$ to $+0.192$: risk identification moves from $+0.376$ to $+0.079$, stress testing from $+0.321$ to $+0.181$, tool-use parameterization by $+0.010$, and the remaining nine subtasks are unchanged. The primary tables retain the original scorer specification to preserve the evaluation protocol used in the main run. We report the later rescoring as a post hoc sensitivity analysis and will release both versions with the harness. These issues motivated the corrected scorer definitions and illustrate why evaluation artifacts need documented assumptions, not just leaderboards.

More broadly, this extends earlier concerns about robustness to representation, data, and distributional variation \citep{sun2025quantifying,sun2022unimodalbias}: in agentic systems, the evaluation interface and execution environment themselves become additional sources of variation.
 
\paragraph{From single-agent to system-level effects.} All results here measure a \emph{single-agent} performance effect: one agent, exogenous prices. In deployment, many agents holding the same packages would trade in the same markets. Because FinSkillBench provides a population of agents with controllable, measured behavioral differences on a shared point-in-time data stack, a natural follow-up varies the share of skill-equipped agents in a multi-agent rebalancing loop with a price-impact model, asking whether the premium survives crowding; we leave this analysis to ongoing work.

\subsection{From Isolated Subtasks to Composed Workflows}
\label{sec:composition}
A desk never runs these subtasks in isolation; it chains them. As a probe of
whether per-subtask skill gains survive composition, we ran a focused case study,
\emph{FinSkillBench-Desk}, that threads the same curated-skill subtasks into a single
nine-stage mandate-delivery chain---mandate parse, fundamental research,
Black--Litterman view, construction, monitoring, stress, risk identification,
remediation, and rebalancing---and scores the final book against a deterministic,
re-derivable theoretical best. Holding the model (\texttt{gpt-4.1}), mandate
(\texttt{medium\_core}), and 30-name universe fixed, we contrast a
\emph{teacher-forced} mode, in which each stage consumes clean oracle inputs, with an
\emph{autonomous} mode, in which each stage consumes the agent's own upstream outputs.

The result is a composition failure that component scores do not see. Teacher-forced,
the chain is fully deployable (100\% mandate adherence, optimality 1.000). Autonomous,
it produces books that are near-optimal in weight space (optimality 0.934) yet never
mandate-adherent (0 of 33), while the entire downstream risk-management control loop
(monitor, stress, risk identification, remediation) still scores 1.000 against its
per-stage conditional oracle (Tables~\ref{tab:desk-terminal} and~\ref{tab:desk-stage}).
The cause is a single seam: stage one parses the mandate into a plausible but
non-canonical contract, the constraint builder silently drops the position, sector,
beta, and tracking-error limits it can no longer find, and the downstream monitor then
certifies compliance against the same degraded contract while its conditional oracle
inherits the identical corruption. We call this \emph{silent competence}: a control
loop that is internally self-consistent and confident while enforcing a mandate it
mis-transcribed upstream. The breach surfaces only at the terminal gate, which is
scored against the immutable canonical mandate rather than the agent's own parse.

A high per-subtask skill score---even the
near-ceiling curated scores of Sec.~\ref{sec:results}---is necessary but not
sufficient for composed deployability, and self-consistent local monitoring is not
assurance. The actionable point is narrow: supervise the contract-parse and
constraint-aware construction handoffs, and drive compliance checks from the canonical
mandate rather than the agent's own parse. Because this is a single-model,
single-mandate probe (Sec.~\ref{sec:limitations}), we read it as an
existence-and-mechanism result that motivates process-level, mandate-adherence
evaluation, not as a prevalence claim.

\begin{table}[ht]
\centering
\small
\caption{FinSkillBench-Desk composed-process probe: terminal outcomes by execution
mode (\texttt{gpt-4.1}, \texttt{medium\_core} mandate, nine-stage chain, $n{=}33$ per
mode). Adherence is judged against the immutable canonical mandate; optimality is
$1-$ normalized weight distance to a deterministic theoretical best. The
teacher-forced versus autonomous separation is categorical, so bootstrap intervals on
adherence are trivially $[1,1]$ and $[0,0]$.}
\label{tab:desk-terminal}
\begin{tabular}{@{}lcc@{}}
\toprule
\textbf{Terminal metric} & \textbf{Teacher-forced} & \textbf{Autonomous} \\
\midrule
Mandate adherence & 1.000 & 0.000 \\
Optimality & 1.000 & 0.934 \\
\bottomrule
\end{tabular}
\end{table}

\begin{table}[ht]
\centering
\small
\caption{FinSkillBench-Desk per-stage local competence: mean score of each stage
against the oracle run on that stage's actual input (headline grid, $n{=}33$ per
mode). Every stage except \texttt{construct} reads $\geq 0.99$ in both modes---
including the entire monitor/stress/risk-identification/remediation loop---yet the
composed book is never mandate-adherent. \texttt{tool\_use} is scored structurally
($0.5$), not as $0/1$ correctness; its causal role is recovered from the run logs
(Sec.~\ref{sec:composition}).}
\label{tab:desk-stage}
\begin{tabular}{@{}lccc@{}}
\toprule
\textbf{Stage} & \textbf{Teacher-forced} & \textbf{Autonomous} & \textbf{TF $-$ Auto} \\
\midrule
\texttt{tool\_use} & 0.500 & 0.500 & 0.000 \\
\texttt{fa\_research} & 1.000 & 1.000 & 0.000 \\
\texttt{view} & 1.000 & 1.000 & 0.000 \\
\textbf{\texttt{construct}} & \textbf{0.751} & \textbf{0.000} & \textbf{0.751} \\
\texttt{monitor} & 1.000 & 1.000 & 0.000 \\
\texttt{stress} & 1.000 & 1.000 & 0.000 \\
\texttt{risk\_id} & 1.000 & 1.000 & 0.000 \\
\texttt{remediate} & 1.000 & 1.000 & 0.000 \\
\texttt{rebalance} & 1.000 & 1.000 & 0.000 \\
\bottomrule
\end{tabular}
\end{table}

\section{Limitations}
\label{sec:limitations}
\paragraph{Trajectory sampling.}
The main experiment uses one trajectory per model, condition, and task cell at
temperature $0.7$, so it estimates resource effects across the task
distribution rather than within cell stochastic success probabilities.
Paired and cluster bootstrap analyses quantify uncertainty across episodes,
but repeated trajectories would provide a more direct estimate of
within agent execution variance.

\paragraph{Benchmark scope.}
FinSkillBench covers a controlled 30 stock U.S. equity universe and three
investment management domains. It does not cover other asset classes,
international markets, derivatives, private assets, or the full range of
institutional workflows. Several targets are generated by deterministic
financial engines and therefore inherit the assumptions of those methods.

\paragraph{Task and resource coupling.}
The curated packages were developed specifically for FinSkillBench.
Accordingly, some measured benefits reflect alignment between task
contracts and the procedural resources supplied. This is most explicit in
risk identification, where the skill document states the output vocabulary
used by the verifier. We therefore interpret that result as a mixture of
procedural and schema guidance rather than pure domain knowledge transfer.

\paragraph{Resource delivery and harness dependence.}
Curated documents are accessed under a 4{,}000 character delivery limit, and
the tools only and curated conditions do not use identical executable tool
interfaces. Hermes provides an independent implementation check, but its
no skill condition also removes reference data carried by domain tools.
These choices bound the causal interpretation of individual resource
contrasts while motivating explicit reporting of the agent harness
configuration.

\paragraph{Composed-workflow probe.}
The composition result (Sec.~\ref{sec:composition}) is a single-model,
single-mandate, single-universe case study (\texttt{gpt-4.1},
\texttt{medium\_core}, 30 names): it establishes the existence and mechanism of
silent competence, not its prevalence across models or difficulty tiers. Its
magnitude is partly co-determined by a fixed tooling confound---a non-canonical
mandate parse and a curated optimizer that cannot express tracking-error or beta
constraints---and the categorical teacher-forced versus autonomous separation
yields degenerate confidence intervals by construction. Prevalence across models
and mandates is ongoing work.
 
\section{Conclusion}

We introduced \textbf{FinSkillBench}, a controlled evaluation suite for
verifiable financial agent workflows with point in time inputs,
regenerable hidden ground truth, and task specific deterministic
verification. Across 9 models and 17{,}820 episodes, curated skill
packages raise mean scores by +16.2 points (0.366 to 0.528) in the
paired analysis across 8 models, while self generated skills add only
+0.5 points. Under the bounded single episode setting studied here,
agent generated procedural documents do not reproduce the benefit of
pre built curated resources. The benefit of a curated package is also
not a single procedural knowledge effect. Documents alone add +5.6
points, executable tools alone add +19.5 points, and their combination
is subadditive. The source of the premium depends on the workflow, with
numerically intensive tasks benefiting primarily from executable
resources, documentation mattering more when procedural or schema level
guidance is the bottleneck, and interpretive workflows benefiting from
both. The pattern is sign stable under 10 scorer variants, and Hermes
Agent reproduced the directional pattern across 8 models with higher
curated means in all three domains, while raw magnitudes vary with
harness tool and data access design. A post hoc audit further shows that
scorer assumptions and execution failures can move headline numbers by
several points.

More broadly, a measured skill premium is not an intrinsic property of the underlying language model. It depends on the resources available to the agent, how those resources are exposed, the interaction budget, and the evaluation contract. Agent benchmarks should therefore treat resource and harness design as part of the scientific object rather than as an implementation detail. A composed extension of the same subtasks into a nine-stage mandate-delivery chain further shows that near-ceiling per-subtask scores need not compose: a curated-skill agent running on its own handoffs delivers a near-optimal but never mandate-adherent book while every per-stage control reports success, underscoring that process-level, canonically-anchored adherence testing---not per-skill accuracy---is what certifies deployability. We will release FinSkillBench, its
skill packages, verifiers, scorers, and evaluation artifacts upon publication to support reproducible study of these system level effects. Natural next steps include matched cross harness effect estimates, interface identical tool arms that separate tool content from access path, and ontology controlled scoring for categorization tasks.


\section*{Ethics and Privacy Statement}
This work uses only public filings, public market data, and author-constructed synthetic data; no personal, client, or proprietary data is used. The benchmark evaluates analytical correctness in controlled tasks, not suitability for autonomous trading or investment advice. Releasing curated skills may lower the barrier to deploying overtrusted financial agents. We mitigate this by providing deterministic verifiers, evaluation outputs, and documented failure modes. To reduce selective reporting, we also release all scorers and the
ten-scheme robustness sweep for independent reproduction and stress testing.

\bibliographystyle{ACM-Reference-Format}
\bibliography{sample-base}

\appendix

\end{document}